\documentclass[conference]{IEEEtran}

\IEEEoverridecommandlockouts                        
\usepackage{times}

\usepackage[numbers]{natbib}
\usepackage{multicol}

\usepackage{hyperref}

\usepackage{xcolor} % for text color
\usepackage{comment} % multiline comment blocks
\usepackage{graphicx}
\graphicspath{ {./images/} }
\usepackage{subfigure}
\usepackage{dblfloatfix} % to hold figures in place

\usepackage[absolute, overlay]{textpos}
\usepackage{etoolbox}

\makeatletter
\patchcmd{\@makecaption}
{\scshape}
{}
{}
{}
\makeatother

\renewcommand{\vec}[1]{\mathbf{#1}} % bold vectors

\newcommand{\anonymize}[1]{RBO~Hand~3}

\title{\LARGE \bf
Surprisingly Robust In-Hand Manipulation: An Empirical Study}

\author{Aditya Bhatt$^\star$ \quad Adrian Sieler$^\star$ \quad Steffen Puhlmann \quad Oliver Brock% <-this % stops a space
\thanks{ $^\star$Both authors contributed equally to this work.}
\thanks{All authors are with the Robotics and Biology Laboratory, Technische
	Universit\"{a}t Berlin, Germany. We gratefully acknowledge financial support by the Deutsche Forschungsgemeinschaft (DFG, German Research Foundation) under Germany’s Excellence Strategy - EXC 2002/1 "Science of Intelligence" - project number 390523135, German Priority Program DFG-SPP 2100 "Soft Material Robotic Systems" - project number 405033880 and DFG project "Robotics-Specific Machine Learning (R-ML)" - project number 329426068.
}}

\textblockorigin{50mm}{5mm} % start everything near the top-left corner
\begin{document}

\begin{textblock}{3}(0,0)
	\noindent
	\centering
	Published in the Proceedings of Robotics: Science and Systems, 2021.
\end{textblock}

\hyphenation{OpenAI}

\maketitle
\thispagestyle{empty}
\pagestyle{empty}

%======================================================================
% Abstract
\begin{abstract}
We present in-hand manipulation skills on a dexterous, compliant, anthropomorphic hand. Even though these skills were derived in a simplistic manner, they exhibit surprising robustness to variations in shape, size, weight, and placement of the manipulated object. They are also very insensitive to variation of execution speeds, ranging from highly dynamic to quasi-static. The robustness of the skills leads to compositional properties that enable extended and robust manipulation programs. To explain the surprising robustness of the in-hand manipulation skills, we performed a detailed, empirical analysis of the skills' performance.  From this analysis, we identify three principles for skill design: 1) Exploiting the hardware's innate ability to drive hard-to-model contact dynamics. 2) Taking actions to constrain these interactions, funneling the system into a narrow set of possibilities. 3) Composing such action sequences into complex manipulation programs. We believe that these principles constitute an important foundation for robust robotic in-hand manipulation, and possibly for manipulation in general.
\end{abstract}
%====================================================================== 

%======================================================================
\section{Introduction}\label{sec:introduction}
%======================================================================

\begin{figure}[h!]
	\centering
	\includegraphics[width=0.98\linewidth]{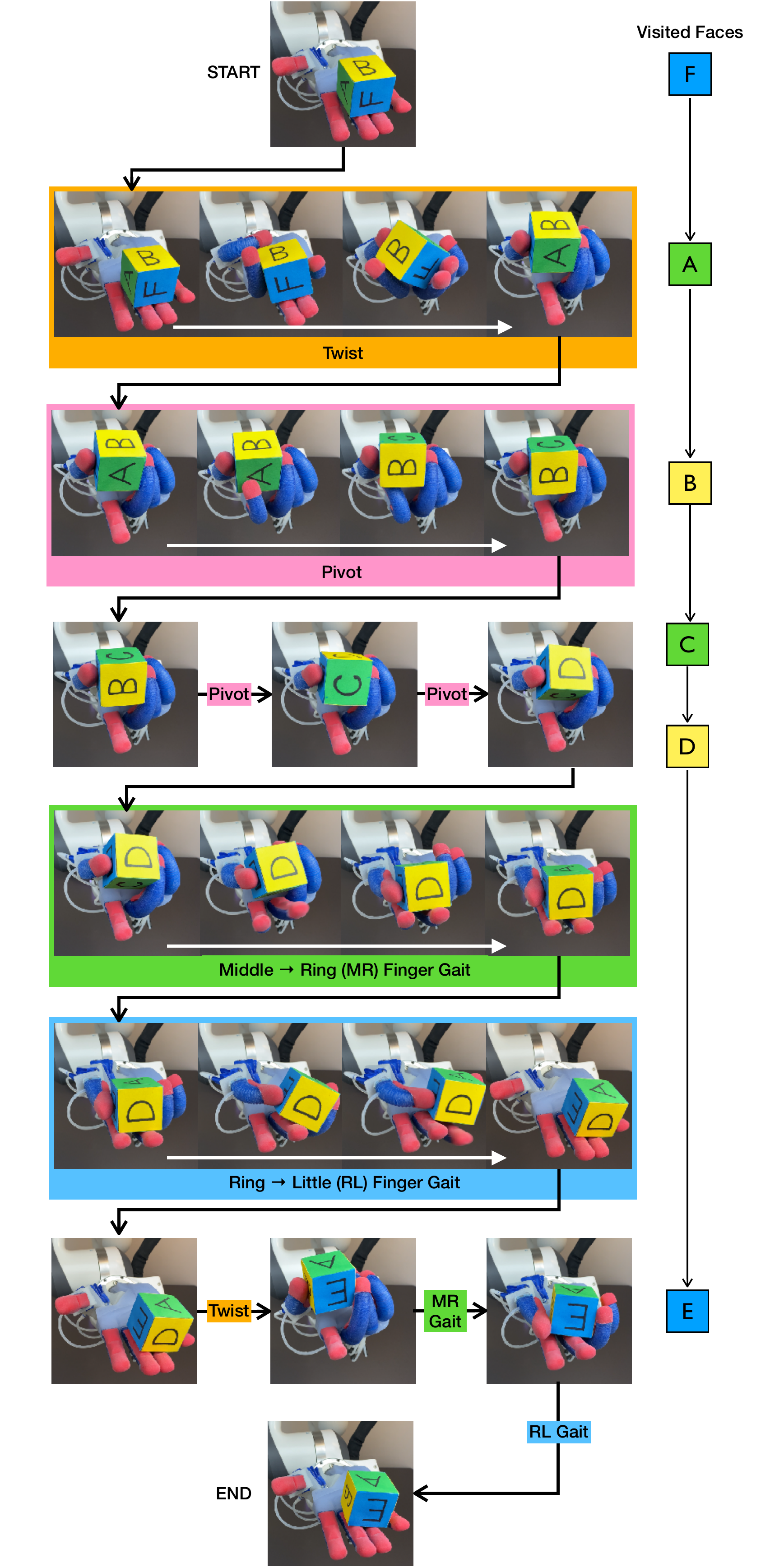}
	\caption{\textbf{An open-loop in-hand manipulation program} that cycles through all cube faces in the order: $FABCDE$. This plan is composed entirely of just four skills executed at appropriate times. With these skills, we can achieve all possible face configurations ($24 = 6$ faces $\times 4$ orientations). The skills feature diverse contact types (sliding, pushing, and rolling) between the object and finger parts (dorsal, ventral, lateral, tip) and feature various capabilities like twisting, pivoting, finger-gaiting, and controlled dropping. Full video: \url{https://youtu.be/jRHIbnE8Kfo}}
	\label{fig:FABCDE}
\end{figure}

We present dexterous in-hand manipulation skills on a robotic hand (\autoref{fig:FABCDE}) that substantially go beyond the state of the art in their robustness, generality, and fluidity of motion.  As it is difficult to substantiate claims about real-world behavior in writing, we suggest that the interested reader watch the following video to form their own opinion: \url{https://youtu.be/Z6ECG3KHibI}.

The state of the art, at this time, is the groundbreaking work presented by OpenAI~\cite{andrychowicz_learning_2020}, who used Deep Reinforcement Learning to produce remarkably dexterous behavior on a five-fingered robotic hand, first manipulating a cube and later even an articulated Rubik's cube~\cite{open_solving_2019}. Their learned skills feature contact-rich movements like finger-gaiting, pivoting, and the exploitation of gravity.

In contrast to OpenAI's work, our approach uses a highly compliant hand and achieves such capabilities entirely without sensing, hand or object models, or machine learning. Yet, we demonstrate in-hand manipulation skills that 
\begin{itemize}
	\item transfer unmodified to objects of diverse shapes, weights, and sizes, e.g.~cubes, cuboids, and cylinders, varying in size and weight by a factor of two,
	\item execute successfully and robustly when their execution speed is varied by a factor of~$80$,
	\item tolerate, without loss of performance, variations in translational object placement of over $90$\% of the object's size and variations in rotational placement of~$23^\circ$, 
	\item can reliably be composed into larger manipulation programs, letting us execute them consecutively up to~$140$ times,
	\item and do all of this with sensorless open-loop actuation.
\end{itemize}
\autoref{fig:FABCDE} shows the execution of such a program, composed of four repeatedly executed in-hand manipulation skills.

We demonstrate the existence of these remarkable generalization phenomena in real-world experiments. We then analyze the resulting behavior to identify, empirically, the reasons for this observed performance. Our key contribution is to illustrate \textit{how} to achieve such robust manipulation behavior. Through our analysis, we hypothesize principles that lead to robust manipulation. 

We believe these insights offer new opportunities for achieving human-level dexterity and facilitate the development of capable robotic manipulation systems. Much like convolution is a useful inductive bias for deep learning in computer vision, the exploitation of these principles may serve as a powerful inductive bias for Deep RL in robotic manipulation.

%All of this is performed without kinematic or dynamics models.

%======================================================================
\section{Designing In-Hand Manipulation Skills}
%======================================================================

In order to study the problem of in-hand manipulation from first principles, we tried the simplest possible approach. We observed ourselves rotating a cube in our human hands, and attempted to replicate this behavior with a robotic hand by manually actuating its parts. This effort resulted in a skill that was easy to design, yet surprisingly advanced in its robustness and manipulation complexity. Before we explain how we designed the skill, we must first introduce the hardware we used, as it plays a salient role. 

%----------------------------------------------------------------------
\subsection{The \anonymize{RBO~Hand~3}}
%----------------------------------------------------------------------

\begin{figure}[tp]
	\centering
	\subfigure{\includegraphics[height=5cm]{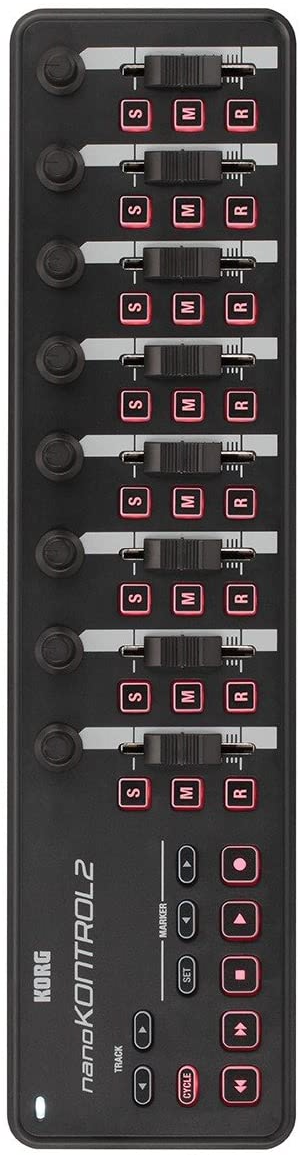}}
	\subfigure{\includegraphics[height=5cm]{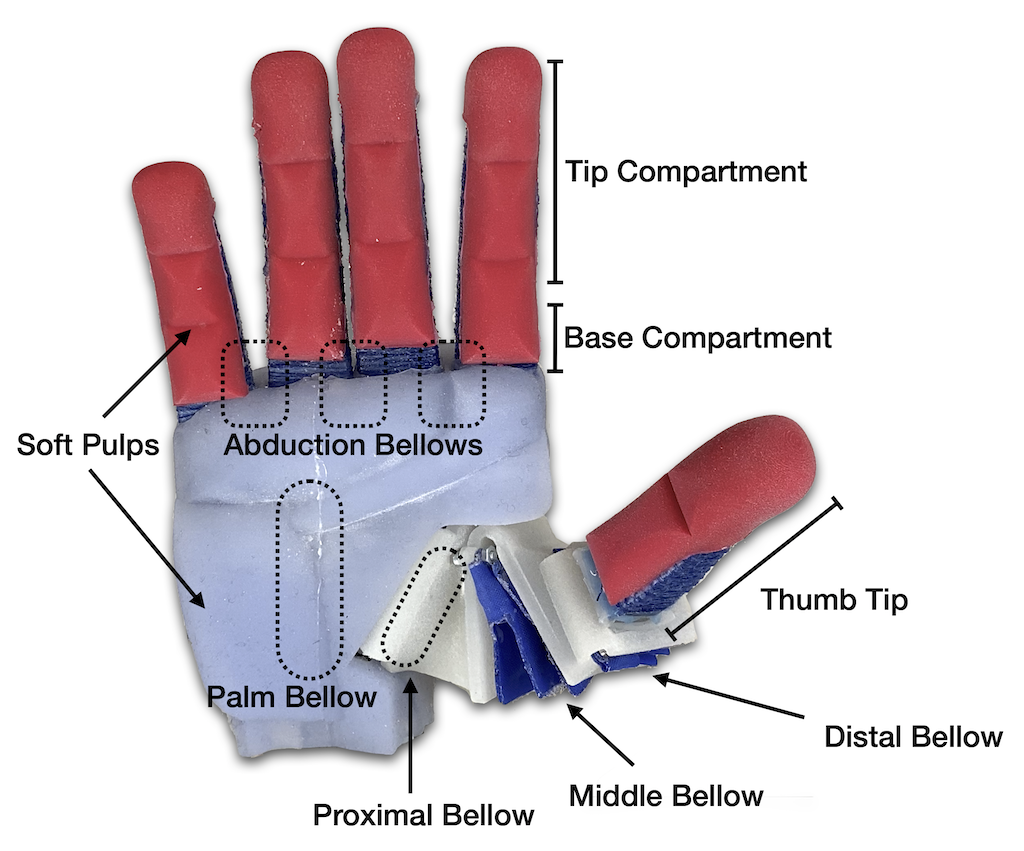}}
	\caption{\textbf{The \anonymize{RBO~Hand~3}}, an anthropomorphic, pneumatically actuated, soft robotic hand with 16 degrees of actuation. It is highly compliant to support the diverse mechanical interactions encountered during manipulation. Soft silicone paddings, called pulps (red and white), cover the fingers and palm. An easy way to manually command hand postures is with a mixing board (pictured left); each slider maps to one actuator.}
	\label{fig:rh3-labeled}
\end{figure}

To explore in-hand manipulation, we used the compliant anthropomorphic \anonymize{RBO~Hand~3}~\cite{RH3} (\autoref{fig:rh3-labeled}). Its silicone fingers are pneumatically actuated, soft continuum actuators with two air chambers each. The hand additinally possesses a highly dexterous opposable thumb, consisting of a stack of differently oriented soft bellow actuators made of coated nylon fabric. Additional bellows inside the hand can spread the fingers wide and flex the palm inwards. In total, the hand has 16 actuators and virtually unlimited degrees of freedom due to its flexible morphology.

We actuate the hand by controlling the air-mass enclosed in each actuator~\cite{air_mass_deimel}. These inflation levels can be commanded by a computer or through a re-tasked electronic mixing board (commonly used in music synthesis).

%----------------------------------------------------------------------
\subsection{Coding In-Hand Manipulation Skills}
%----------------------------------------------------------------------

The first behavior we chose to replicate, after observing our own hands, was to repeatedly rotate a cube counter-clockwise. To faithfully mimic this behavior on the robot hand, we did not restrict ourselves to the common practice of fingertip-only contacts. As we shall see later, this is crucial to the success of our approach.

We placed the cube at a specific location on the area spanned by the ring and middle fingers. Using a mixing board, we manually adjusted the actuators' inflations and observed the resulting hand-object interactions. With this process, we were able to produce a similar counter-clockwise rotation. 

Next, we captured these commanded inflation levels into a sequence of \textit{keyframes}, corresponding to the enclosed air-masses in all actuators $\vec{a} = \{a^1, a^2, \dots, a^{16} \}$. We then replayed this sequence onto the hand by linearly interpolating through the intermediate keyframes.

\begin{figure}
	\centering
	\includegraphics[width=0.7\linewidth]{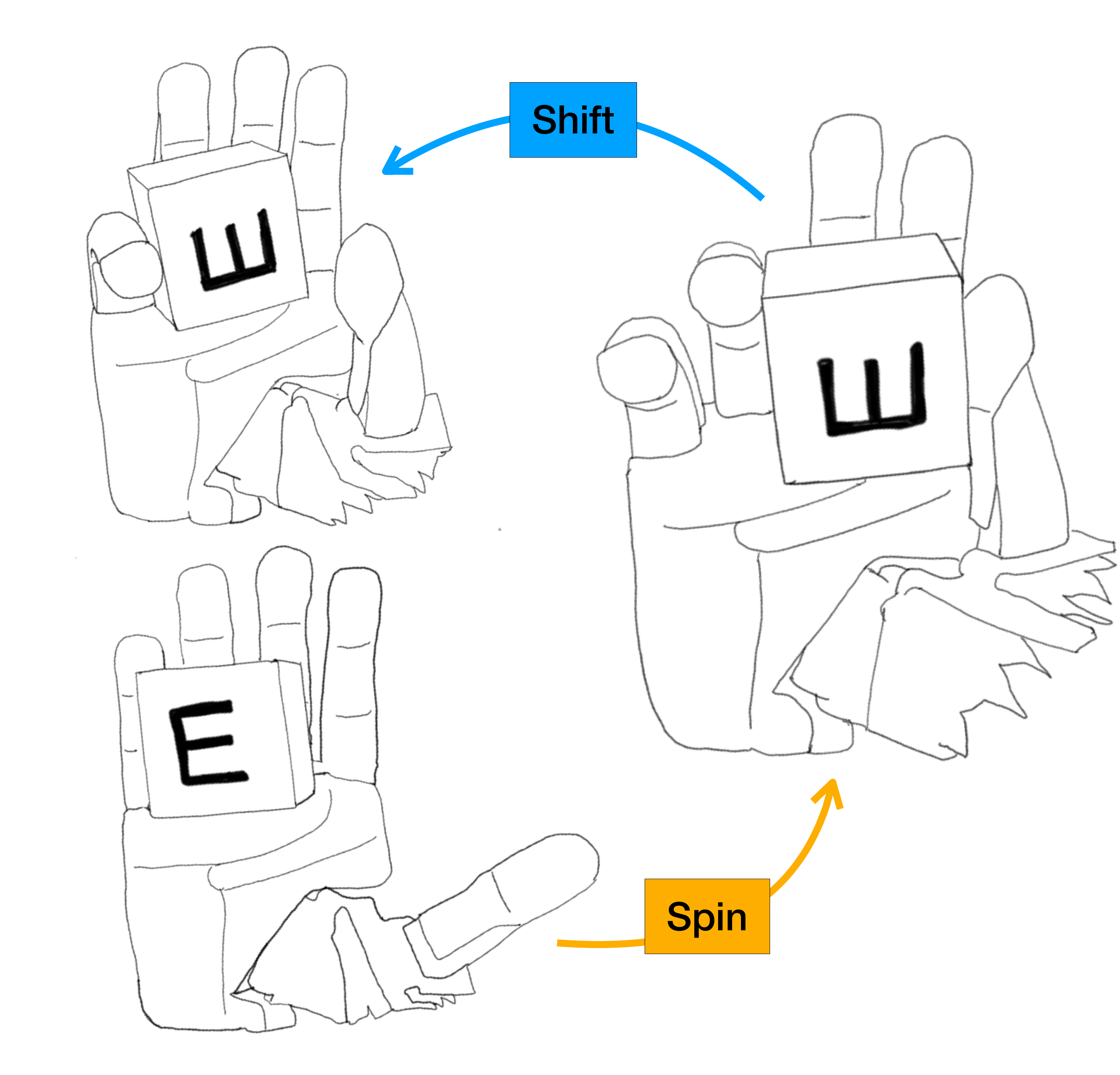}
	\caption{\textbf{Start and end states of the \textit{spin} and \textit{shift} skills}. We replicated this behavior on the robotic hand by observing ourselves perform the same behavior.}
	\label{fig:spinshift-doodle}
\end{figure}

\begin{figure*}[h]
	\centering
	\includegraphics[width=1\linewidth]{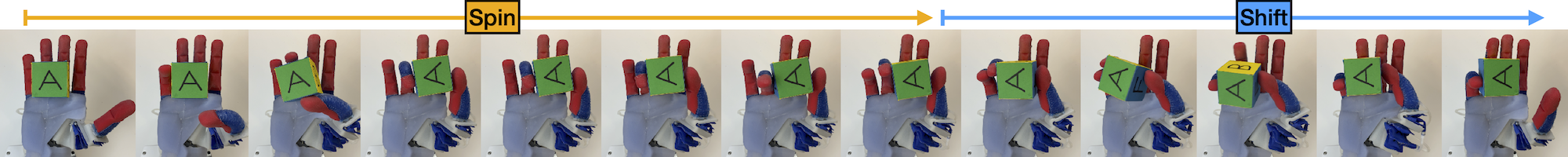}
	\caption{\textbf{All keyframes of the designed \textit{spin} and \textit{shift} skills} (Viewed from above): \textit{Spin} uses the thumb and ring finger to rotate an object counter-clockwise. \textit{Shift} gaits the cube from the ring finger to the little finger, moving it to the left.}
	\label{fig:keyframes-spinshift}
\end{figure*}

\begin{figure}[]
	\centering
	\includegraphics[width=1\linewidth]{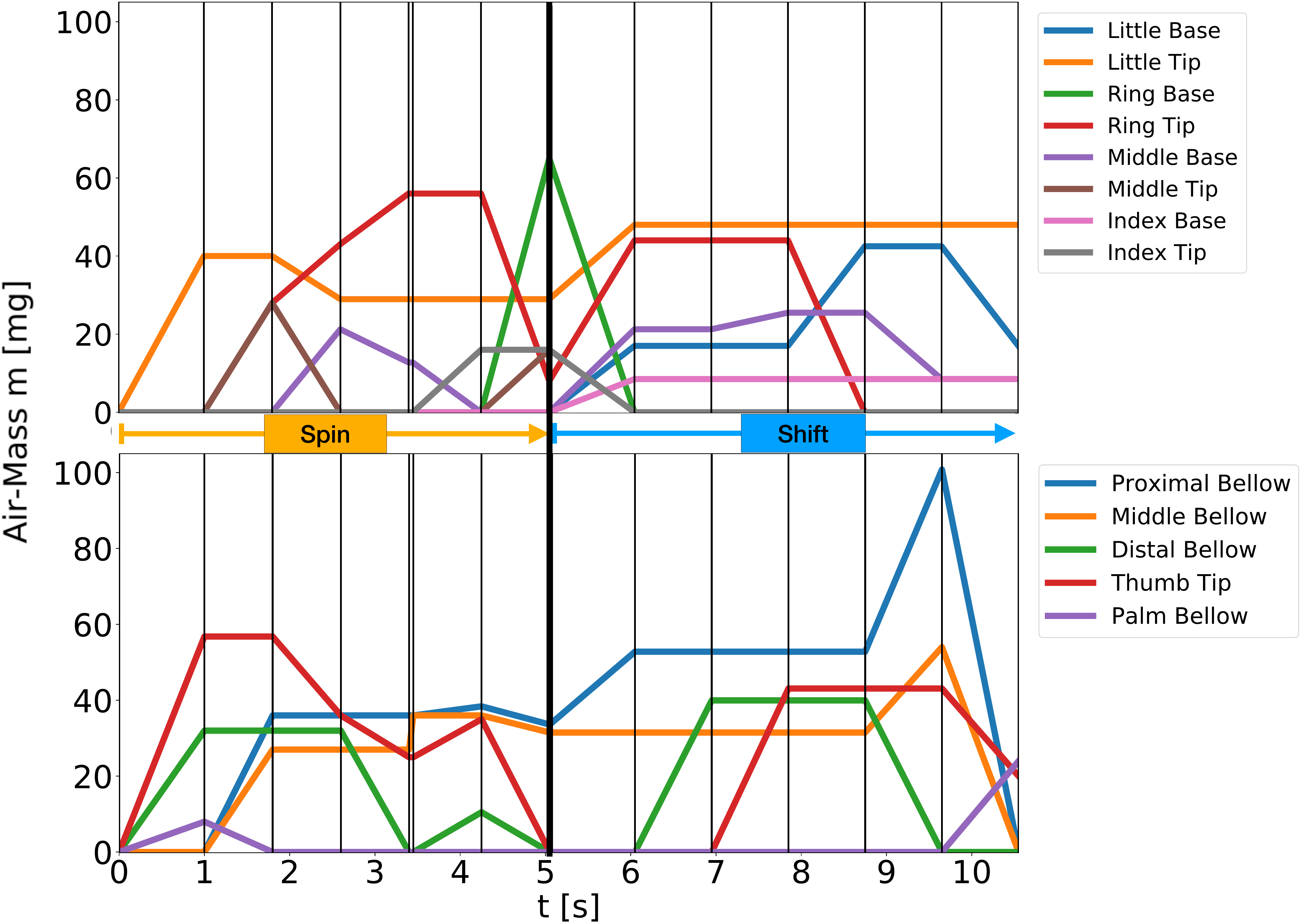}
	
	\caption{\textbf{Air-mass actuation signal for the \textit{spin + shift} skill.} Each vertical line corresponds to an intermediate keyframe. Appendix \ref{sec:appendix} contains a detailed description of each of these keyframes.}
	\label{fig:air_mass_spin_shift}
\end{figure}

Without the visual feedback we used when actuating the hand interactively, we expected the rotation to fail during automated replay.  Instead, it worked! With additional fine-tuning of the keyframes, we were able to reliably spin the cube by $90^\circ$ each time and (with another sequence of keyframes) shift it back to its starting location as illustrated in \autoref{fig:spinshift-doodle}. Designed in this manner, the \textit{spin} and \textit{shift} skills are open-loop interpolations through a total of $13$ hand-crafted keyframes (\autoref{fig:air_mass_spin_shift}), depicted visually in \autoref{fig:keyframes-spinshift}. In the same way, we designed three additional skills (depicted in \autoref{fig:FABCDE}) that work together to reorient a cube to all possible face configurations. We summarize all of these:
\begin{itemize}
	\item \textbf{Spin} uses the thumb to rotate the cube ounter-clockwise by $90^\circ$ around the ring-finger, and places it to rest on the index and middle finger. It does so with 7 keyframes.
	\item \textbf{Shift}, also called \textbf{Ring$\rightarrow$Little (RL) Finger Gait}, gaits the cube from the ring-finger to the little finger, and places it to rest on the ring and middle finger (5 keyframes).
	\item \textbf{Twist} uses the thumb, middle, and ring fingers to lift the cube into a precision grip, in the process rotating the cube counter-clockwise by $90^\circ$ (9 keyframes).
	\item \textbf{Pivot} maintains this posture, contacting the cube with the index finger to rotate it by $90^\circ$ around the grip axis \\(6 keyframes).
	\item \textbf{Middle$\rightarrow$Ring (MR) Finger Gait}, a variant of \textit{shift}, gaits the cube from this posture to a grasp between the thumb and middle finger, and places it to rest on the index and ring fingers (6 keyframes).
\end{itemize}
The corresponding air-mass actuation sequences for these particular three skills can be found in the Appendix \ref{sec:appendix}.

The main point of this section is to illustrate the simplicity of the programmed in-hand manipulation skills: they are open-loop sequences of keyframes, programmed using human intuition. One might call this approach simplistic.  Still, the resulting skills are highly robust.  In the next section, we present and analyze experimental evidence for this robustness.

%======================================================================
\section{Evidence of Surprising Robustness}
\label{sec:evidence_robustness_spin_shift}
%======================================================================

We now experimentally characterize the robustness of the hand-programmed skills. We report results for the \textit{spin + shift} skill, as well as for the combined \textit{twist + pivot + MR-gait + shift}\footnote{The \textit{RL-gait (shift)} skill is a sightly updated version of the one used in \textit{spin + shift}. Nevertheless, all manipulations referenced in the subsequent videos for the combined \textit{twist + pivot + MR-gait + shift} skill are performed with the exact same open-loop actuation signal.} skill. The sequencing of these particular skills allows us to restore the manipulated object --- if executed successfuly --- to a configuration that can be handled by the \textit{twist} as well as the \textit{spin} skill.

%----------------------------------------------------------------------
\subsection{Robustness to Object Placement}
%----------------------------------------------------------------------

Due to its sensorless design, it is reasonable to expect the skill to only work for a specifically chosen initial cube pose. However, this is not the case. \url{https://youtu.be/-gzeFdHbvIM} shows successful and unsuccessful trials of \textit{spin + shift} with different initial placements. Surprisingly, the skill works robustly for a wide set of reasonable poses. We can identify regions of the cube's pose space which would lead to failure. The \textit{twist + pivot + MR-gait + shift} skill is similarly robust (video: \url{https://youtu.be/gzvRXcdolpc}). In the next section, we will investigate why.

%----------------------------------------------------------------------
\subsection{Generalizing to New Objects}
%----------------------------------------------------------------------

\begin{figure}[h!]
	\centering
	\includegraphics[width=1\linewidth]{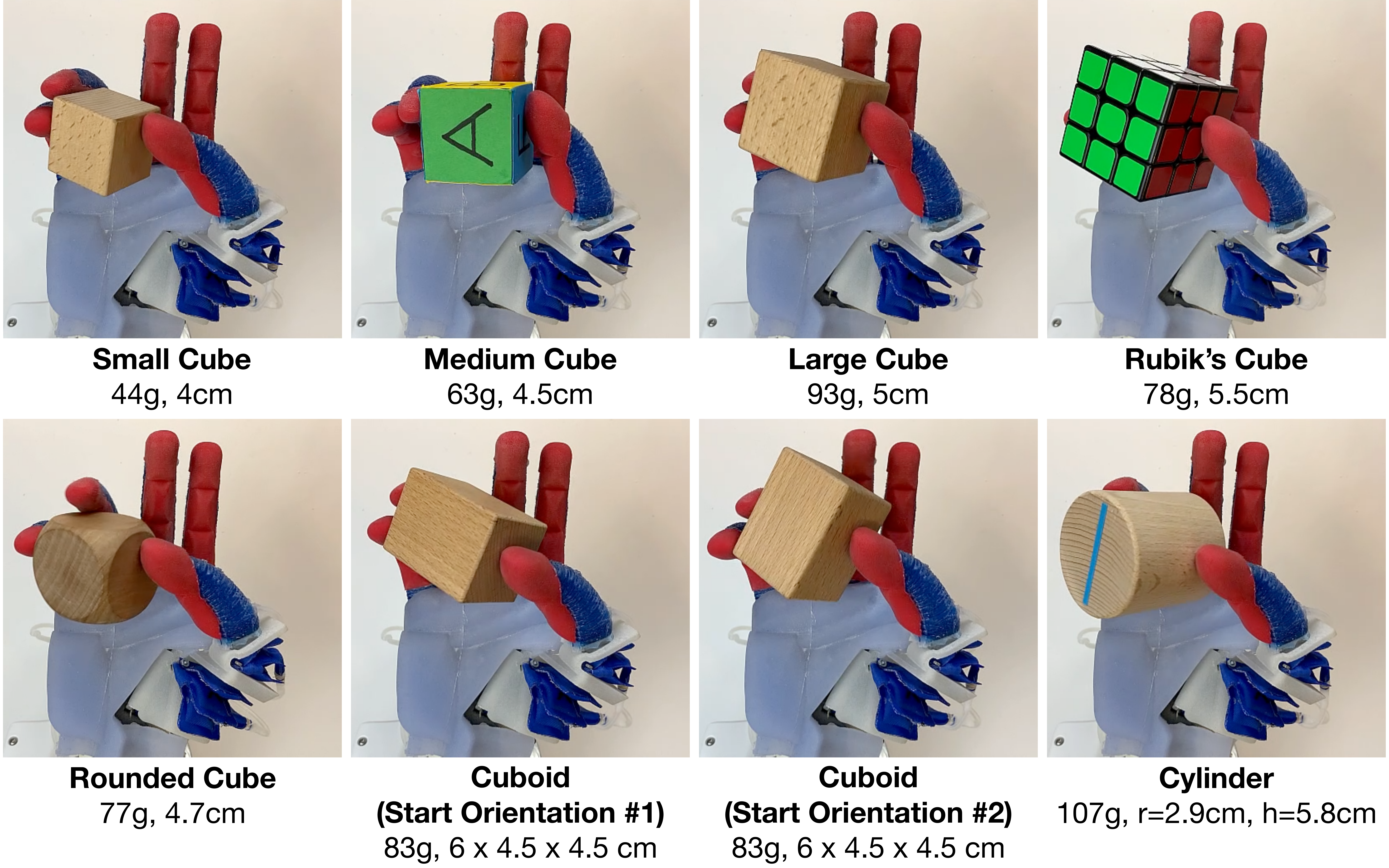} 
	\caption{\textbf{\textit{Spin + shift} successfully manipulating seven objects of varying shape, size, weight, and surface properties, using the same open-loop actuation signal.} The sizes of the cubes and the cuboid refer to the edge length(s). Cylinder - $d$ represents the diameter and $h$ the height.}
	\label{fig:spin_shift_objects}
\end{figure}

\autoref{fig:spin_shift_objects} and its corresponding video (\url{https://youtu.be/TejM-kfllVE}) show the same \textit{spin + shift} skill executing successfully with seven different objects: variously sized cubes, a cuboid, and even a cylinder. Having designed the skill using only two different cubes, we were surprised by this immediate generalization across shapes, sizes, and weights. This observation is not restricted to the \textit{spin + shift} skill. The \textit{twist + pivot + MR-gait + shift} skill exhibits similar transfer across objects (video: \url{https://youtu.be/NNf5UBgQQZk}).

The actuation signal is the same in all cases, so clearly our skills cannot manipulate every possible object (e.g. \textit{spin + shift} fails to rotate pyramids or spheres), but the straightforward generalization we observe makes detailed knowledge of objects' properties unnecessary.

Why, then, do we see this object-adaptive behavior? We will discuss this in a later section.

%----------------------------------------------------------------------
\subsection{Robustness to Execution Speed}
%----------------------------------------------------------------------

As designed, \textit{spin + shift} executes for approximately $11$ seconds. However, this duration is arbitrary. We recorded executions with a range of different speeds; a video (\url{https://youtu.be/JU6BlpaqjVA}) shows the results. The slowest executions we tested ran for over $1$ minute, while the fastest reliable runs took only $0.8$ seconds, suggesting a temporal scaling tolerance of at least $80$. This degree of robustness to timing changes was surprising: the skill had only ever been tested at the same speed of about $11$ seconds, and the dynamics of the manipulandum and the relative influence of gravity vary significantly over these different time scales. The more complex \textit{twist + pivot + MR-gait + shift} skill was designed to execute over a duration of $21$ seconds. We found it robust to execution time-scales as low as $7.5$ seconds, and as high as $5.5$ minutes. (video: \url{https://youtu.be/-T6DMiI6veE}).

%----------------------------------------------------------------------
\subsection{Composing Longer Manipulations}
%----------------------------------------------------------------------

The \textit{spin + shift} skill leaves the object in approximately the same position at which it starts. Having observed the skill's robustness to variability in initial positions, we looped its execution. In our demonstration (video: \url{https://youtu.be/6OAWxvhpfdQ}), we were able to spin a Rubik's cube $140$ times before it falls out due to an unexpected slippage. The fact that this is possible without any kind of explicit feedback control makes it all the more surprising. Again, this observation is not limited to the \textit{spin + shift} skill. \url{https://youtu.be/jJr3UxtbFCs} shows the \textit{twist + pivot + MR-gait + shift} skill repeatedly manipulating a $4.5$cm cube $54$ consecutive times.

%======================================================================
\section{Empirical Analysis of Robustness}
\label{sec:empirical_study}
%======================================================================

Having established that these skills are highly robust, we now attempt to identify the reasons for this robustness by empirical analysis.  Again, due to space constraints, and the impracticality of motion-capture for the other skills, we only analyze the \textit{spin + shift} skill. The analyses of the other skills should lead to identical conclusions.

%----------------------------------------------------------------------
\subsection{Robustness During Manipulation}
%----------------------------------------------------------------------

\begin{figure}[h!]
	\centering
	\includegraphics[width=1\linewidth]{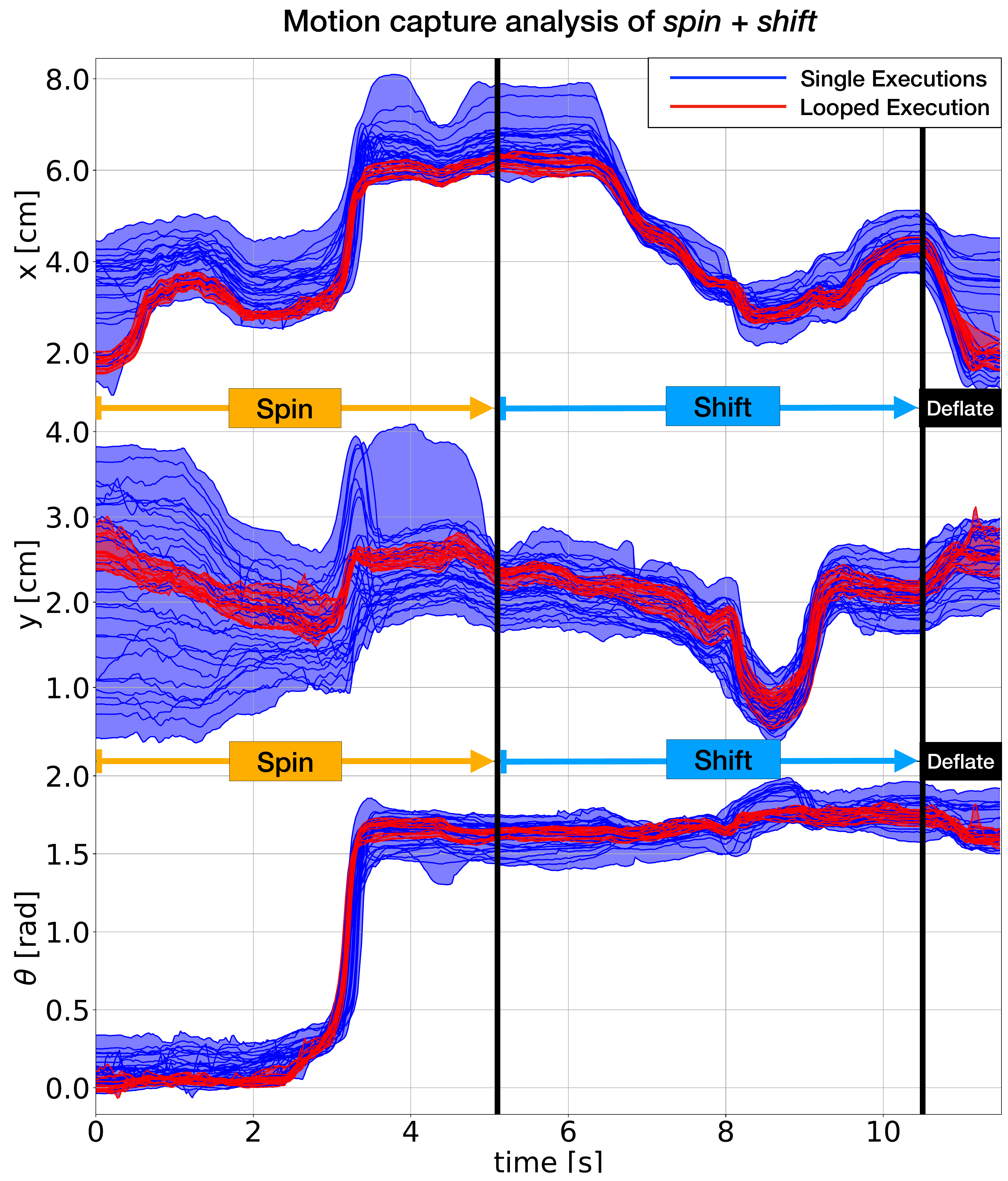}
	\caption{\textbf{Quantitative evidence of robustifying disturbance rejection}, through tracking of the $(x, y, \theta)$ pose of a wooden cube, while executing the \textit{spin + shift} skill. The blue curves are $33$ independent trials. The $15$ red curves are traced by a single episode of uninterrupted looped execution. After each execution episode, we deflate all compartments to rule out any drift in the low-level air-mass controller.}
	\label{fig:mocap-envelopes}
\end{figure}

\begin{figure*}[h!]
	\centering
	\includegraphics[width=1\linewidth]{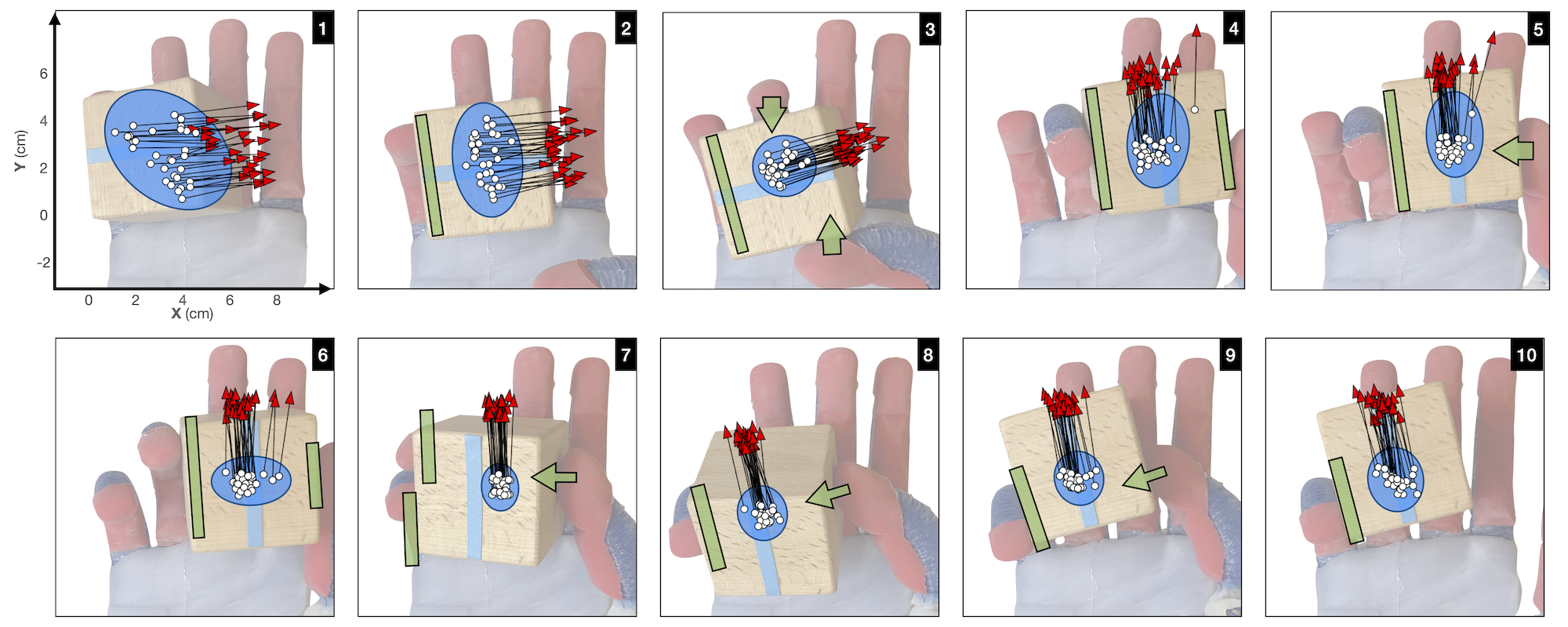}
	\caption{\textbf{Robust manipulation as a series of uncertainty-restricting constraint exploitations:} The plots depict a keyframe-by-keyframe breakdown of cube poses $(x,y,\theta)$ gathered over 33 independent trials of \textit{spin + shift}. The underlaid photographs are sourced from only the most salient keyframes of one illustrative execution. A white dot marks the cube position from a single execution, and the associated arrow indicates the planar rotation $\theta$. The blue region coarsely indicates the set of observed cube positions over all trials. The green bars represent physical walls implemented by the fingers, and the green arrows represent active pushing interactions.
	}
	\label{fig:mocap-constraints}
\end{figure*}

To understand the high reliability of our sequenceable skills, we must look closely at how these keyframe transitions affect the object. We placed the hand in a 16-camera motion capture system and tracked a $4.5$cm wooden cube's pose in the $(x,y)$ plane coincident with the palm. The outstretched fingers point along the $y$-axis (like in \autoref{fig:mocap-constraints}).

As our manipulation skills are open-loop, the most straightforward way to quantify their robustness is to execute the skill for different initial $(x,y)$ positions, thereby characterizing the region of success. \autoref{fig:mocap-envelopes} shows all 33 successful trajectories of the
\textit{spin + shift} skill in blue. Outside of the observed initial range of $3.5$cm, $4.5$cm, and $23^\circ$ along $x$, $y$, and $\theta$, the manipulation fails. This indicates a tolerance margin of over 90\% of the cube's own width along the $y$ axis.

More interesting than the skill's ``range of robust applicability'' is \textit{why} the manipulation is robust. Within the envelope of successful trials, we see these independent blue trajectories being funneled from a wide range of initial positions into a dense inner trough as the manipulation executes.  This suggests that the manipulation achieves robustness by actively restricting the cube's position to a narrow range of possibilities.

We can test this idea by repeatedly looping the same skill, without intervention, to see if the trajectories remain restricted. \autoref{fig:mocap-envelopes} shows these 15 looped executions (plotted in red) over the blue envelope. Despite unintended slippages, imprecise contact locations, and sensorless actuation, the cube keeps orbiting in a narrow basin of attraction without deviating. 

We conclude that the robustness is somehow achieved via attractor dynamics that reject disturbances. Using the same trajectory data, we will now study \textit{how} this robustifying uncertainty-restriction works, keyframe-by-keyframe.

%----------------------------------------------------------------------
\subsection{The Building Blocks of a Robust Skill}
\label{sec:anatomy_of_a_manipulation_skill}
%----------------------------------------------------------------------

Using the same trajectory data, we now study \textit{how} this robustifying uncertainty-restriction works at the level of each keyframe. Each plot in \autoref{fig:mocap-constraints} depicts the $(x,y,\theta)$ pose of the cube after commanding a keyframe. We now describe what happens in the keyframes, referring to the figure.

The cubes start out with a large scattering in initial positions. In keyframe 2 (KF2), the hand raises the little finger to ensure the cube's placement near the middle-ring area.  This is a robust operation that restricts the cube's horizontal scattering. In KF3, the ring finger and thumb tips act as a constraint to actively clamp the cube from either side, now squashing their vertical scattering. Clearly, this cannot succeed if the thumb cannot move behind the cube as shown. The skill would fail if the cube were too large or positioned entirely on the palm. It would also fail if the cube were placed too close to the fingertips.

So far, we have inspected translational manipulations. During KF3$\rightarrow$KF4, we strongly curl the ring finger inwards while uncurling the thumb. This greatly increases the staggered opposing forces, causing a rapid rotation that could increase uncertainty. However, in KF4 the cube' poses remain well-restricted over the index-middle area, since the thumb and ring finger now act as constraining walls to the left and right of the cube. Additionally, the ring finger's presence shifts the cube \textit{exactly one finger-width} to the right. The rapid motion still \textit{increases} the uncertainty vertically, as there are no guards in that direction, other than friction.  

In KF5, the thumb moves, pushing the cube against the curved ring finger. This aligns them, robustly squashing rotational uncertainty. This is an instance of a well-understood class of parts-orienting sensorless movements, previously studied in domains other than in-hand manipulation~\cite{goldberg_orienting_1993}.

The remaining keyframes belong to the \textit{shift} skill; they apply a sequence of the same push-to-align operations. These constraints, while enforced by different fingers at different times, remain similar. The constraints are also mobile, due to the fingers' intrinsic lateral compliance. Using them, \textit{shift} performs a guarded translation by one finger-width to the left.

We see that \textit{spin+shift} is robust because it is itself composed of a series of highly robust primitive manipulations. These primitives work well, because they constrain the set of possible outcomes using physical constraints, produced and carefully positioned by the hand.

%----------------------------------------------------------------------
\subsection{Limits of the Spin and Shift Movement}
%----------------------------------------------------------------------

Having investigated the robustness of our skill, we now discuss the ways in which it can fail. A collection of failure cases can be found in this video: (\url{https://youtu.be/roLVk5pKQX8})

To successfully trigger a rotation in the \textit{spin} skill, the object must first be firmly clamped between the thumb and ring-finger, as in KF3 (\autoref{fig:mocap-constraints}). The skill cannot be executed if this precondition is not satisfied, which could happen for the following reasons.

\noindent
\textbf{Positional and Orientational Limits}
\begin{itemize}
	\item The skill begins executing with the object sitting on the little finger, because raising this finger would push the object into the palm.
	\item The object starts out resting very close to the fingertips, making it difficult for the ring finger to curl and guard the cube from falling out.
	\item The object rests mainly on the palmar region so the thumb cannot move behind it.
	\item The object is placed close to the index finger, in which case the thumb is unable to block the object from falling off the right side.
	\item The object is not oriented in a way that allows the thumb and ring finger to make contact on opposing sides. Therefore, the spin transition cannot be executed.
\end{itemize}
\noindent
\textbf{Shape and Size limits}
\begin{itemize}
	\item The object is too big, obstructing finger motion, preventing the establishment of constraints
	\item The object is too small for the thumb to make contact
	\item The object is too heavy, bending the compliant fingers, changing the constraints provided by them
\end{itemize}
Similar limits apply to the \textit{shift} skill. A collection of failures due to initial object placement for the \textit{twist + pivot + MR-gait + shift} skill can be found in this video: (\url{https://youtu.be/w2-NNi3Nnro}).

%======================================================================
\section{Principles for Robust Manipulation}
%======================================================================

Having studied these skills' inner workings, we now discuss our findings and identify three design principles for achieving dexterous in-hand manipulation. We will see that these principles are not new, in fact, they have been known for decades. We will thus also speculate on why the progress reported here became possible only now. The subsequent section will then discuss the related work that is also based on the three principles identified below.

%----------------------------------------------------------------------
\subsection{Outsourcing Control of Contact Dynamics}
\label{sec:outsourcing}
%----------------------------------------------------------------------

We exploit the deeply researched insight that compliant mechanisms can free us from several challenging computational problems~\cite{passive_dynamic_walker, brown_universal_2010}. We describe two of these ways below.

The most straightforward advantage of compliant hardware is its tendency to adapt to an object's shape, due to which simple open-loop control (e.g. closing all fingers together) is sufficient to grasp unknown objects; an ability that is nontrivial to achieve with fully actuated hands~\cite{bicchi_modelling_2011, catalano_adaptive_2014, deimel_novel_2016, santina_toward_2018}.
In this paper, we see the effect of shape adaptation in \autoref{fig:spin_shift_objects}, where \textit{spin + shift} manipulates 7 different objects of various shapes, sizes, and weights using the same actuation signal. We therefore conclude that compliant shape adaptation is useful not only in grasping, but also in fine manipulation.  

Less obviously, soft compliant hardware frees us from the unfulfillable need to accurately model contact interactions. We explain this with an example. 

\begin{figure}[h!]
	\centering
	\includegraphics[width=1\linewidth]{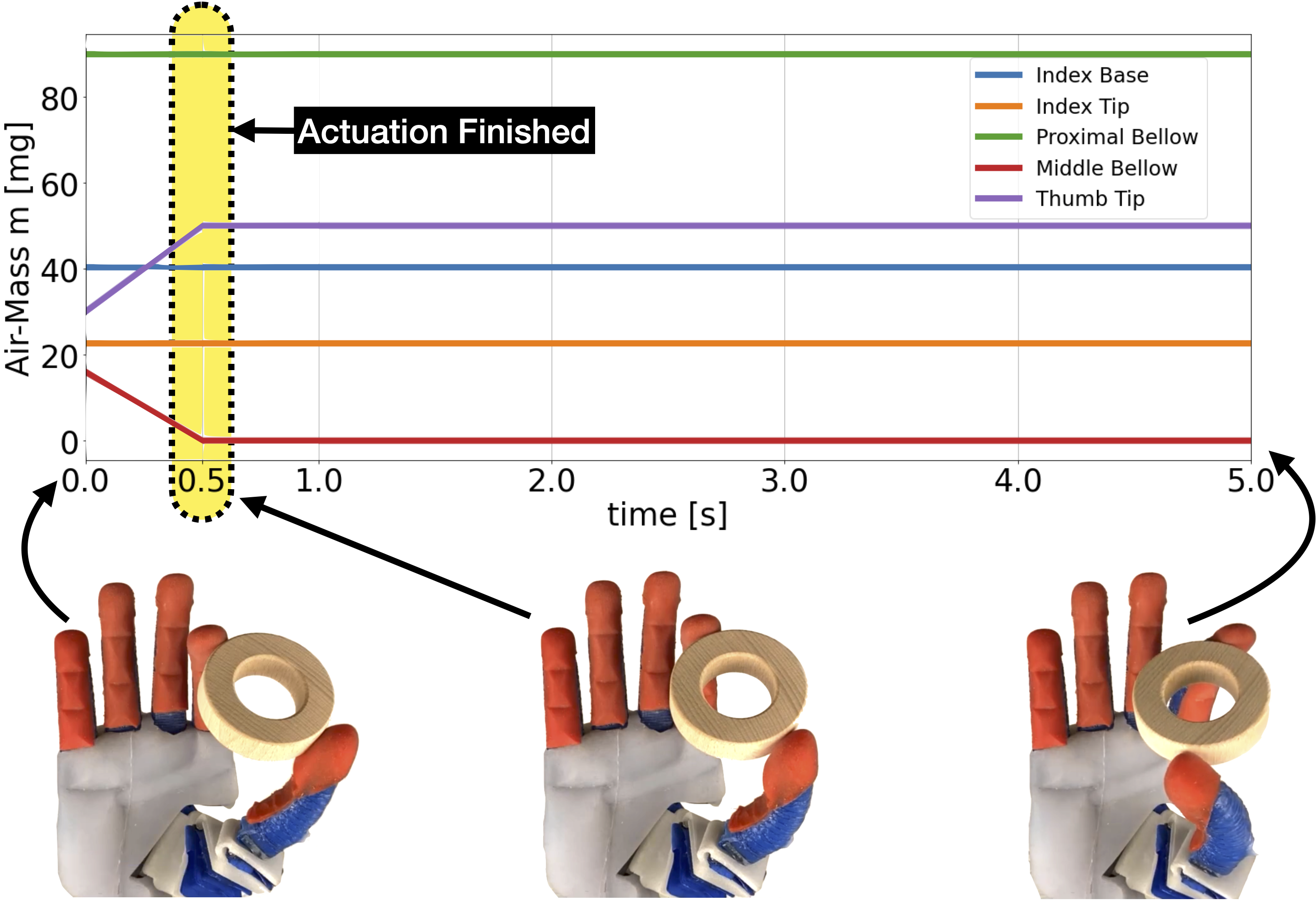}
	\caption{\textbf{The hand twists a wooden donut in a stable rolling pinch grasp, without any feedback control.} Between 0 and 0.5 seconds, we actuate the thumb tip and the distal bellow. The actuation is finished long before the majority of the manipulation unfolds, showing the temporally-extended nature of the invoked dynamic response of the hand. This interaction is fully reversible by restoring the original inflation levels.}
	\label{fig:rh3-rolling}
\end{figure}

\autoref{fig:rh3-rolling} shows a within-grasp manipulation with a rolling contact (best seen in a video: \url{https://youtu.be/Nor2QEtM4W8}) This manipulation consists of just two keyframes. In the first, we hold a wooden donut in a precision grip between the thumb and the index finger. With actuation, the donut twists stably without losing contact. Despite actuating only the thumb mechanism, we see that the index finger moves significantly, remaining diametrically opposed to the thumb. Its motion is \textit{solely} due to mechanical compliance and contact friction, which in this case is crucial for the motion's stability.

Note that the manipulation of the donut continues long after the actuation has completed at $0.5s$. We argue that the hardware itself executes an object-adaptive \textit{closed-loop} manipulation primitive. We never compute this explicitly; the process plays out entirely as a result of the hand's morphology, in the sense of \textit{morphological computation}~\cite{paul_mc_2006, ghazi-zahedi_morphological_2017}. The demonstrated contact dynamics are prohibitively complex to model analytically, but nevertheless perform very robustly in the real world. This manipulation behavior also transfers to different objects, and can be executed in reverse, as shown in this video: \url{https://youtu.be/6yiSLq59yFA}. Notably, the speed of motion varies for differently sized objects, and also at different intermediate postures.

This approach is closely related to another well-known control framework. Our air-mass actuation scheme resembles commanding a \textit{virtual pose} in the sense of impedance control~\cite{hogan_impedance_1985}. In impedance control, linear spring-like behavior encoded in a control law generates interaction forces whenever the hand/fingers deviate from their free-motion poses~\cite{hogan_impedance_1985_2}. In our case, we do not design a control law but merely expect the actuator's elastic properties to evoke such a response. For example, the amount of air enclosed in a finger uniquely determines its virtual pose as well as a stiffness profile. On the \anonymize{RBO~Hand~3} platform, the combination of 16 actuators lets us cycle through a vast number of virtual poses, each of which recruits different dynamic responses of the hand when in contact with an object. This computation-free alternative to managing mechanical interactions was suggested three decades ago by Hogan~\cite{hogan_impedance_1985}:

\textit{``Exploiting the intrinsic properties of mechanical hardware can also provide a simple, effective and reliable way of dealing with mechanical interaction."}

This forms our first design principle: we can identify and exploit useful and robust hand-object interactions, and compose them into dexterous manipulations. We lose the ability to command arbitrary interaction dynamics. What we gain in exchange is a greatly simplified control problem, with several substantial benefits, as explained above.

%----------------------------------------------------------------------
\subsection{Constraining Object Motion}
%----------------------------------------------------------------------

The dynamic response of the hand can produce complex behaviors when invoked by actuation, but at the expense of precise object motion. It is difficult to guarantee, for example, a specific object pose as a result of the interaction. In the absence of such certainty, we can instead restrict how the dynamics unfold. This idea is central to the study of \textit{constraint exploitation}~\cite{perez, mason_sensorless}, and has been used to ease manipulation with the aid of pre-existing environmental constraints such as walls, corners, and fixtures~\cite{eppner_exploitation_2015, shao_learning_2020}. In our skills, we also use constraints as tools to make manipulation work, but these are not external environmental features.

We explain this with the \textit{spin + shift} example from \autoref{fig:mocap-constraints}. During the KF3$\rightarrow$KF4 transition, we trigger a very dynamic interaction: a rapid spin of the cube. The reason this transition does not spin the cube more than $90^\circ$, or send it flying out of the hand, is that KF4 blocks the cube with the ring finger and thumb on either side. Constraining the dynamics of the manipulandum in this way makes the resulting behavior highly regular, assuring a specific outcome with great certainty.

Each keyframe transition in our skills is robust manipulation in its own right. All of these primitives offload the fine details of interactions to the hand's compliant morphology, and recruit physical constraints from that same morphology to prevent undesirable outcomes.

This forms our second design principle: the morphology itself serves as a platform for establishing \textit{constraints} on the dynamics of the hand/object interaction. Under this framework, the skills we can design are limited by the available constraints, and we lose the ability to command arbitrary object poses. In practice, this need not be a problem; by actuating a sufficiently complex morphology, we can generate a combinatorially large number of possible constraints, letting us robustly achieve diverse object configurations.

%----------------------------------------------------------------------
\subsection{Composing Manipulation Funnels}
\label{sec:funnels}
%----------------------------------------------------------------------

As \autoref{fig:mocap-constraints} shows, our open-loop primitives restrict uncertainty in object location by geometrically squashing, stretching, and moving the set of possible placements. Restricting uncertainty through mechanical action is the very idea of \textit{manipulation funnels}, pioneered by Mason~\cite{mason_mechanics_1985} over three decades ago. We quote from the paper: \\
\textit{``Using a funnel, the goal to position an object can be accomplished despite variation in the initial locations and shapes of the objects."}

As our primitives also restrict uncertainty in this way, we regard our primitives as funnels. The set of states from which a funnel can be executed is called its \textit{entrance}. Upon execution, it is guaranteed to map these states to another set called the \textit{exit}. The power of funnels lies in their potential for composition~\cite{mason_sensorless, empirical_backprojections, burridge_funnels_1999}. If the exit of one funnel lies in the entrance of another, then by executing both funnels, we are guaranteed to end up in the second funnel's exit. Importantly, such a composition is also a funnel in its own right.

This constitutes our third design principle: by sequencing several of these primitive funnels, we can design long complex manipulation funnels like \textit{spin} and \textit{shift}. Some funnels have exits that lie in their own entrances, e.g. \textit{spin+shift} as well as \textit{twist + pivot + MR-gait + shift}; we can loop their execution indefinitely. 

\begin{figure}[h!]
	\centering
	\includegraphics[width=1\linewidth]{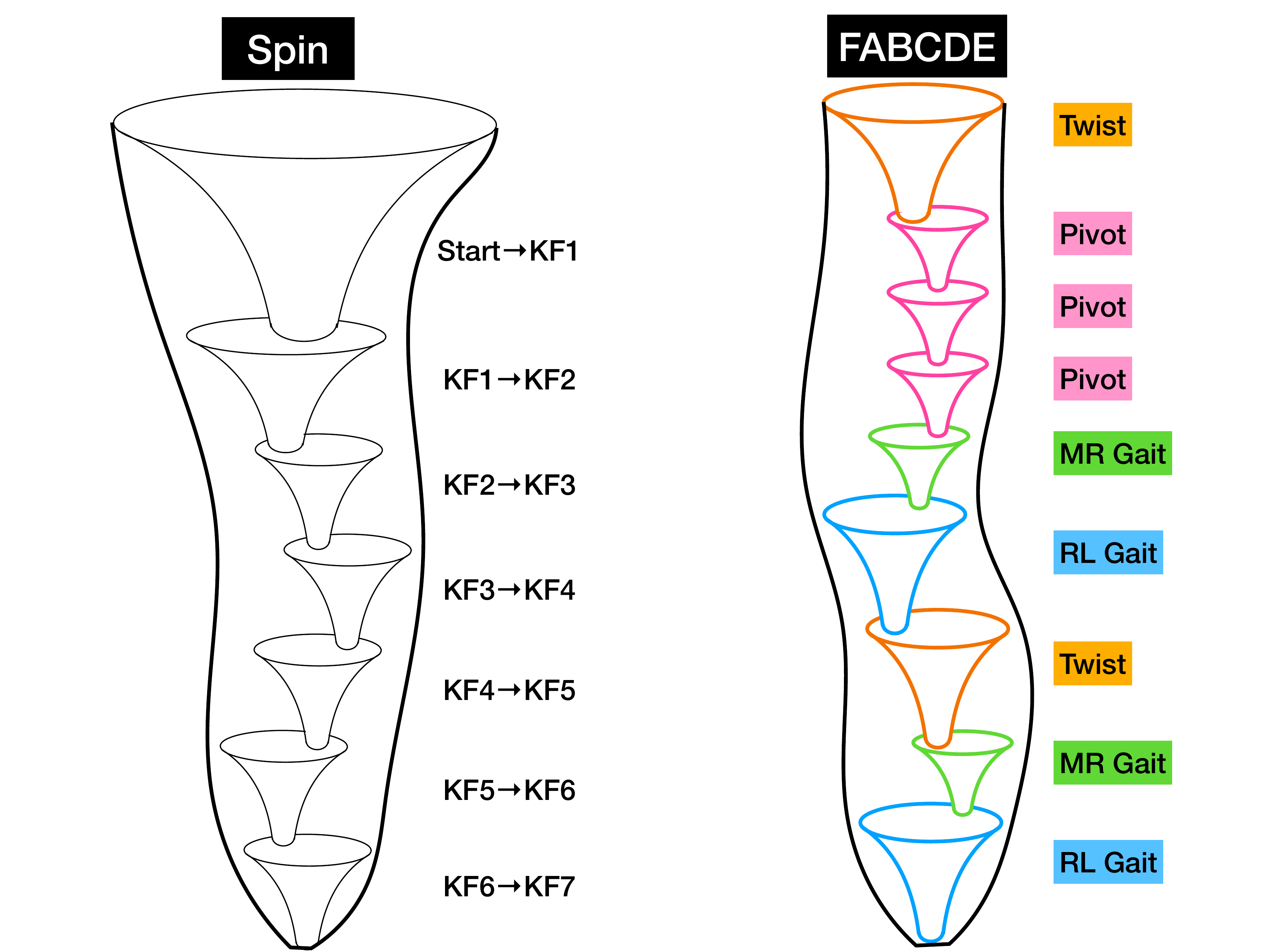}
	\caption{\textbf{Robust manipulation through funnel composition:} Like \textit{spin}, each skill is composed of funnel-like robust primitive manipulations, so it is a robust funnel in itself. Each funnel reliably transforms a set of hand-object configurations (its entrance) into another set (its exit). We designed our skills to funnel into each other, letting us compose longer manipulation plans like FABCDE. } 
	\label{fig:funnel-composition}
\end{figure}

We have designed all of our dexterous skills as funnel compositions, letting us incorporate them into long manipulation plans like the cube reorientation program in \autoref{fig:FABCDE}. \autoref{fig:funnel-composition} informally depicts the funnel composition used in the program. 

In our final demonstration, we rely on the cross-object generalization of our chainable skills to manipulate multiple lettered objects in one open-loop execution, to spell a word (video: \url{https://youtu.be/tnq0xXMUbhc}). This demo combines sequences of all five discussed skills, as well as open-loop grasping and placement skills involving  a 7-DoF Franka Emika Panda robot arm. The Appendix \ref{sec:appendix} contains a detailed description of the funnels used in this demonstration.

%----------------------------------------------------------------------
\subsection{Why Now?}
\label{sec:why_now}
%----------------------------------------------------------------------

The three principles identified above are known for decades, begging the question why the progress presented here was not accomplished decades ago. We believe there are several reasons for this.

While the principles were known in their respective disciplines, they were not identified as \textit{the} key ingredients. As a result, few researchers made it their focus to combine them into a single approach. Instead, our community debated and explored the advantages and limitations of analytical approaches versus data-driven approaches. In retrospect, analytical approaches mostly ignored the fact that we may not need accurate models for robust manipulation. (Note, however, that our insights do not call into question the value of those analytical models.)  And data-driven approaches usually tackled problems that were much more complicated than necessary, because they largely ignored the inductive biases provided by the three principles.

The dependence on capable hardware posed another obstacle for progress in the past. In this study, we demonstrate that the most suitable realization of the three principles requires the careful co-design of hand morphology and associated control strategies~\cite{co_design}. Only when the implementation of the three design principles is orchestrated across both hardware and software, can it achieve its full potential. The advent and success of soft robotics as a research discipline has produced many insights necessary for the design of hardware sufficiently competent for the full realization of the three principles~\cite{RH3}. Centrally among them is the ability of soft hands to establish safe contact over large surface areas, as opposed to the point-based contact models underlying most analytic approaches to manipulation.

The explanations offered above can be illustrated by returning to the work of OpenAI, mentioned in the beginning of this paper~\cite{open_solving_2019,andrychowicz_learning_2020}. The work pursued a strongly data-centric approach, based on domain randomization. This approach forwent the advantages of outsourcing the handling of contact dynamics to clever hardware, did not leverage the inductive bias represented by the exploitation of physical constraints provided by the hand, and did not explicitly incorporate a compositionality prior \textit{à la} funnels. In some sense, it attempted something much harder than what the problem required. We are excited to see the power of data-driven approaches and, in particular, deep reinforcement learning joined with the inductive biases presented here.

%======================================================================
\section{Related Work}
%======================================================================

The principles we described here are prevalent in past manipulation research. It is impossible to do justice to the many outstanding works that implemented them. Still, we structure our discussion of related work based on these principles. We also discuss research on analytical and data-driven approaches, as we feel these are two important categories of ongoing manipulation research.

%----------------------------------------------------------------------
\subsection{Outsourcing Control of Contact Dynamics}
%----------------------------------------------------------------------
Appropriately designed hardware can simplify the control problem for interaction tasks that require handling of difficult contact dynamics. Two prominent examples are the passive dynamic walker~\cite{passive_dynamic_walker} and the universal robotic gripper~\cite{brown_universal_2010}.
Outsourcing aspects of control and perception to hardware is at the very core of the growing field of soft robotics~\cite{materials_science,sot_robotics_rus}. In the domain of robotic manipulation compliant hands like the Pisa/IIT SoftHand 2~\cite{santina_toward_2018}, the RBO~Hand~2~\cite{deimel_novel_2016}, the \mbox{i-Hy} Hand~\cite{odhner_compliant_2014} as well as the dexterous gripper presented in~\cite{abondance_dexterous_2020} have shown to be well suited for grasping and in-hand manipulation. Besides hands, the dynamic properties of a system composed of a soft arm with a soft gripper can be used to robustly solve various interaction tasks~\cite{soft_arm_gripper}.

%----------------------------------------------------------------------
\subsection{Constraining Object Motion}
%----------------------------------------------------------------------
Deliberate contacts with the environment can be utilized to constrain the motion of the object to be manipulated. By following this principle, even simple gripper-like systems can attain comprehensive manipulation skills~\cite{dafle_extrinsic_2014,Rodriguez_Prehensile}. Humans employ similar strategies when grasping. The observed environmental constraints can be used to execute robust grasping strategies with a soft robotic hand~\cite{eppner_exploitation_2015}. Environmental constraints can also simplify the learning of peg-insertion tasks~\cite{shao_learning_2020}. The concept of constraint exploitation has been researched for in-hand manipulation with an underactuated gripper. This time the end-effector itself provides the constraints to the objects motion~\cite{ma_toward_2017}.

%----------------------------------------------------------------------
\subsection{Using Funnels and Their Compositionality}
\label{sec:funnel_compositionality}
%----------------------------------------------------------------------

A funnel, as a primitive to reduce uncertainty, is a well-suited abstraction to solve various tasks. The composition of funnels follows a hybrid nature; executing a funnel represents the continous aspect while switching between funnels is by definition a discrete operation. The roots of funnels for manipulation planning can be traced back to Mason's \textit{Manipulation Funnel}~\cite{mason_mechanics_1985} and pre-image back chaining~\cite{perez}. Robust motion strategies for orienting planar objects~\cite{mason_sensorless} and pushing~\cite{mason_push} that reduce uncertainty purely through mechanical interaction are an effective way of outsourcing computation if the task mechanics can be described. This approach has been complemented to obtain the respective motion stratgies from observational data~\cite{empirical_backprojections}. Mason's funnel metaphor and the pre-image back chaining by Lozano-P\'{e}rez, Mason, and Taylor were extended to also cover sensory feedback to switch between funnels and/or feedback-controllers that can act robustly in the presence of uncertainties.  Examples include paddle juggling~\cite{burridge_funnels_1999}, move-to-grasp~\cite{controller_refinement}, real-time motion planning~\cite{funnel_tedrake}, and robot grasping~\cite{eppner_planning}.

%----------------------------------------------------------------------
\subsection{Analytic Approaches}
%----------------------------------------------------------------------

The main platforms for in-hand manipulation research in the last decades have been fully-actuated~\cite{grebenstein_hand_2012} or hyper-actuated robot hands~\cite{butterfass_dlr-hand_2001} with tightly integrated sensory devices, where every degree of freedom is controllable. Different analytic approaches to manipulation using these systems are, for example, impedance control~\cite{pfanne_object-level_2020}, trajectory optimization~\cite{sundaralingam_relaxed-rigidity_2019}, and hybrid position/force-control~\cite{michelman_forming_1994,or_position-force_2016}. Most of the approaches depend on explict models of contact, mostly point contacts, which is not always a realistic assumption in real-world applications.

%----------------------------------------------------------------------
\subsection{Data-Driven Approaches}
%----------------------------------------------------------------------

We discussed the data-driven approach by OpenAI in \autoref{sec:why_now}. Data-driven approaches to grasping~\cite{data_driven_grasp} and in-hand manipulation~\cite{rajeswaran_learning_2018,nagabandi_deep_2019} have gained substantial popularity in recent years.  They often do not make constraining assumptions about contact types or models, thus holding the promise of delivering more general manipulation behavior. These approaches often require massive computational resources and produce strategies encoded in black-box representations. Learning-based methods can complement the capabilities of analytic approaches, e.g.~to obtain impedance gains for grasping and within-grasp manipulation~\cite{li_learning_2014}, to hierarchically structure a cube manipulation problem~\cite{veiga_hierarchical_2020}, or to leverage dynamic movement primitives for within-grasp manipulation~\cite{solak_learning_2019}.

%====================================================================== 
\section{Conclusion}
%======================================================================

We attempted to create the simplest possible in-hand manipulation skill on the \anonymize{RBO~Hand~3}, as a starting point for further investigations. What we thought to be the simplest approach, also turned out to be surprisingly competent and robust.  So robust, in fact, that we think the skills presented in this paper substantially advance the state of the art in in-hand manipulation. In the paper, we substantiate this claim with extensive experimental evidence. This evidence shows how the skills transfer unmodified to objects of diverse shapes, weights, and sizes. It shows that the execution speed of the identical skill can be changed 80-fold without loss in performance, that the skills are robust to substantial disruptions in initial conditions, and that they can be composed into complex in-hand manipulation programs. Our surprise about these results is amplified by the fact that none of the skills we describe here depends on explicit sensing or feedback control.

The near-accidental discovery of very robust in-hand manipulation skills triggered an extensive empirical investigation into the causes of the surprising robustness. Through this investigation, we identified three well-known principles as the underlying pillars of robust in-hand manipulation (and probably of robust manipulation in general). First, it is important to outsource the control of complex and difficult-to-model contact dynamics to the hand's morphology.  Second, the exploitation of constraints to limit the manipulandum's motion leads to significant reduction of uncertainty and thus to simplification of perception and control. Such constraints can arise from the hand's morphology itself or can be offered by the environment.  Third, by viewing the exploitation of constraints as manipulation funnels, we can exploit their compositionality to produce complex manipulation programs. We believe that the insights presented in this paper offer new opportunities for robotic manipulation systems, in the quest of achieving human-level dexterity.

%======================================================================
%\section*{Appendix}
\appendix
The supplementary material can be found at this URL:
\label{sec:appendix}

%======================================================================
\url{http://dx.doi.org/10.14279/depositonce-12071}

%%%%%%%%%%%%%%%%%%%%%%%%%%%%%%%%%%%%%%%%%%%%%%%%%%%%%%%%%%%%%%%%%%%%%%%%%%%%%%%%

%% Use plainnat to work nicely with natbib. 

\bibliographystyle{IEEEtran}
\bibliography{paper.bib}

\begin{thebibliography}{10}
\providecommand{\url}[1]{#1}
\csname url@samestyle\endcsname
\providecommand{\newblock}{\relax}
\providecommand{\bibinfo}[2]{#2}
\providecommand{\BIBentrySTDinterwordspacing}{\spaceskip=0pt\relax}
\providecommand{\BIBentryALTinterwordstretchfactor}{4}
\providecommand{\BIBentryALTinterwordspacing}{\spaceskip=\fontdimen2\font plus
\BIBentryALTinterwordstretchfactor\fontdimen3\font minus
  \fontdimen4\font\relax}
\providecommand{\BIBforeignlanguage}[2]{{%
\expandafter\ifx\csname l@#1\endcsname\relax
\typeout{** WARNING: IEEEtran.bst: No hyphenation pattern has been}%
\typeout{** loaded for the language `#1'. Using the pattern for}%
\typeout{** the default language instead.}%
\else
\language=\csname l@#1\endcsname
\fi
#2}}
\providecommand{\BIBdecl}{\relax}
\BIBdecl

\bibitem{andrychowicz_learning_2020}
M.~Andrychowicz, B.~Baker, M.~Chociej, R.~Józefowicz, B.~McGrew, J.~Pachocki,
  A.~Petron, M.~Plappert, G.~Powell, A.~Ray, J.~Schneider, S.~Sidor, J.~Tobin,
  P.~Welinder, L.~Weng, and W.~Zaremba, ``Learning dexterous in-hand
  manipulation,'' \emph{The International Journal of Robotics Research},
  vol.~39, no.~1, pp. 3--20, 2020.

\bibitem{open_solving_2019}
\BIBentryALTinterwordspacing
I.~Akkaya, M.~Andrychowicz, M.~Chociej, M.~Litwin, B.~McGrew, A.~Petron,
  A.~Paino, M.~Plappert, G.~Powell, R.~Ribas, J.~Schneider, N.~Tezak,
  J.~Tworek, P.~Welinder, L.~Weng, Q.~Yuan, W.~Zaremba, and L.~Zhang, ``Solving
  rubik's cube with a robot hand,'' 2019. [Online]. Available:
  \url{https://arxiv.org/abs/1910.07113}
\BIBentrySTDinterwordspacing

\bibitem{RH3}
S.~Puhlmann, J.~Harris, and O.~Brock, ``{RBO Hand} 3 -- a platform for soft
  dexterous manipulation,'' 2022.

\bibitem{air_mass_deimel}
R.~{Deimel}, M.~{Radke}, and O.~{Brock}, ``Mass control of pneumatic soft
  continuum actuators with commodity components,'' in \emph{IEEE/RSJ
  International Conference on Intelligent Robots and Systems (IROS)}, 2016, pp.
  774--779.

\bibitem{goldberg_orienting_1993}
K.~Y. Goldberg, ``Orienting polygonal parts without sensors,''
  \emph{Algorithmica}, vol.~10, no.~2, pp. 201--225, 1993.

\bibitem{passive_dynamic_walker}
R.~{Tedrake}, T.~W. {Zhang}, {Ming-fai Fong}, and H.~S. {Seung}, ``Actuating a
  simple 3d passive dynamic walker,'' in \emph{IEEE International Conference on
  Robotics and Automation (ICRA)}, vol.~5, 2004, pp. 4656--4661.

\bibitem{brown_universal_2010}
E.~{Brown}, N.~{Rodenberg}, J.~{Amend}, A.~{Mozeika}, E.~{Steltz}, M.~R.
  {Zakin}, H.~{Lipson}, and H.~M. {Jaeger}, ``Universal robotic gripper based
  on the jamming of granular material,'' \emph{Proceedings of the National
  Academy of Sciences}, vol. 107, no.~44, pp. 18\,809--18\,814, 2010.

\bibitem{bicchi_modelling_2011}
A.~Bicchi, M.~Gabiccini, and M.~Santello, ``Modelling natural and artificial
  hands with synergies,'' \emph{Philosophical Transactions of the Royal Society
  B: Biological Sciences}, vol. 366, no. 1581, pp. 3153--3161, 2011.

\bibitem{catalano_adaptive_2014}
M.~G. Catalano, G.~Grioli, E.~Farnioli, A.~Serio, C.~Piazza, and A.~Bicchi,
  ``Adaptive synergies for the design and control of the pisa/{IIT}
  {SoftHand},'' \emph{The International Journal of Robotics Research}, vol.~33,
  no.~5, pp. 768--782, 2014.

\bibitem{deimel_novel_2016}
R.~{Deimel} and O.~{Brock}, ``A novel type of compliant and underactuated
  robotic hand for dexterous grasping,'' \emph{The International Journal of
  Robotics Research}, vol.~35, no.~1, pp. 161--185, 2016.

\bibitem{santina_toward_2018}
C.~D. {Santina}, C.~{Piazza}, G.~{Grioli}, M.~G. {Catalano}, and A.~{Bicchi},
  ``Toward dexterous manipulation with augmented adaptive synergies: The
  pisa/{IIT} {SoftHand} 2,'' \emph{{IEEE} Transactions on Robotics}, vol.~34,
  no.~5, pp. 1141--1156, 2018.

\bibitem{paul_mc_2006}
C.~Paul, ``Morphological computation: A basis for the analysis of morphology
  and control requirements,'' \emph{Robotics and Autonomous Systems}, vol.~54,
  no.~8, pp. 619--630, 2006.

\bibitem{ghazi-zahedi_morphological_2017}
K.~Ghazi-Zahedi, R.~Deimel, G.~Montúfar, V.~Wall, and O.~Brock,
  ``Morphological computation: The good, the bad, and the ugly,'' in
  \emph{{IEEE}/{RSJ} International Conference on Intelligent Robots and Systems
  ({IROS})}, 2017, pp. 464--469.

\bibitem{hogan_impedance_1985}
N.~{Hogan}, ``Impedance control: An approach to manipulation: Part 1 —
  theory,'' \emph{Journal of Dynamic Systems, Measurement, and Control}, vol.
  107, no.~1, pp. 1--7, 1985.

\bibitem{hogan_impedance_1985_2}
------, ``Impedance control: An approach to manipulation: Part 2 —
  implementation,'' \emph{Journal of Dynamic Systems, Measurement, and
  Control}, vol. 107, no.~1, pp. 8--16, 1985.

\bibitem{perez}
T.~{Lozano-Pérez}, M.~T. {Mason}, and R.~H. {Taylor}, ``Automatic synthesis of
  fine-motion strategies for robots,'' \emph{The International Journal of
  Robotics Research}, vol.~3, no.~1, pp. 3--24, 1984.

\bibitem{mason_sensorless}
M.~A. {Erdmann} and M.~T. {Mason}, ``An exploration of sensorless
  manipulation,'' \emph{IEEE Journal on Robotics and Automation}, vol.~4,
  no.~4, pp. 369--379, 1988.

\bibitem{eppner_exploitation_2015}
C.~{Eppner}, R.~{Deimel}, J.~{Álvarez-Ruiz}, M.~{Maertens}, and O.~{Brock},
  ``Exploitation of environmental constraints in human and robotic grasping,''
  \emph{The International Journal of Robotics Research}, vol.~34, no.~7, pp.
  1021--1038, 2015.

\bibitem{shao_learning_2020}
\BIBentryALTinterwordspacing
L.~{Shao}, T.~{Migimatsu}, and J.~{Bohg}, ``Learning to scaffold the
  development of robotic manipulation skills,'' \emph{arXiv}, 2019. [Online].
  Available: \url{http://arxiv.org/abs/1911.00969}
\BIBentrySTDinterwordspacing

\bibitem{mason_mechanics_1985}
M.~T. {Mason}, ``The mechanics of manipulation,'' in \emph{International
  Conference on Robotics and Automation (ICRA)}, vol.~2, 1985, pp. 544--548.

\bibitem{empirical_backprojections}
A.~D. {Christiansen}, ``Manipulation planning for empirical backprojections,''
  in \emph{IEEE International Conference on Robotics and Automation
  (ICRA)}.\hskip 1em plus 0.5em minus 0.4em\relax {IEEE} Computer Society,
  1991, pp. 762--768.

\bibitem{burridge_funnels_1999}
R.~R. Burridge, A.~A. Rizzi, and D.~E. Koditschek, ``Sequential composition of
  dynamically dexterous robot behaviors,'' \emph{The International Journal of
  Robotics Research}, vol.~18, no.~6, pp. 534--555, 1999.

\bibitem{co_design}
R.~{Deimel}, P.~{Irmisch}, V.~{Wall}, and O.~{Brock}, ``Automated co-design of
  soft hand morphology and control strategy for grasping,'' in \emph{IEEE/RSJ
  International Conference on Intelligent Robots and Systems (IROS)}, 2017, pp.
  1213--1218.

\bibitem{materials_science}
M.~{McEvoy} and N.~{Correll}, ``Materials science. materials that couple
  sensing, actuation, computation, and communication,'' \emph{Science}, vol.
  347, p. 1261689, 2015.

\bibitem{sot_robotics_rus}
D.~{Rus} and M.~{Tolley}, ``Design, fabrication and control of soft robots,''
  \emph{Nature}, vol. 521, pp. 467--75, 2015.

\bibitem{odhner_compliant_2014}
L.~U. {Odhner}, L.~P. {Jentoft}, M.~R. {Claffee}, N.~{Corson}, Y.~{Tenzer},
  R.~R. {Ma}, M.~{Buehler}, R.~{Kohout}, R.~D. {Howe}, and A.~M. {Dollar}, ``A
  compliant, underactuated hand for robust manipulation,'' \emph{The
  International Journal of Robotics Research}, vol.~33, no.~5, pp. 736--752,
  2014.

\bibitem{abondance_dexterous_2020}
S.~{Abondance}, C.~B. {Teeple}, and R.~J. {Wood}, ``A dexterous soft robotic
  hand for delicate in-hand manipulation,'' \emph{{IEEE} Robotics and
  Automation Letters}, vol.~5, no.~4, pp. 5502--5509, 2020.

\bibitem{soft_arm_gripper}
H.~{Jiang}, Z.~{Wang}, Y.~{Jin}, X.~{Chen}, P.~{Li}, Y.~{Gan}, S.~{Lin}, and
  X.~{Chen}, ``Hierarchical control of soft manipulators towards unstructured
  interactions,'' \emph{The International Journal of Robotics Research}, 2021.

\bibitem{dafle_extrinsic_2014}
N.~C. {Dafle}, A.~{Rodriguez}, R.~{Paolini}, B.~{Tang}, S.~S. {Srinivasa},
  M.~{Erdmann}, M.~T. {Mason}, I.~{Lundberg}, H.~{Staab}, and T.~{Fuhlbrigge},
  ``Extrinsic dexterity: In-hand manipulation with external forces,'' in
  \emph{IEEE International Conference on Robotics and Automation (ICRA)}, 2014,
  pp. 1578--1585.

\bibitem{Rodriguez_Prehensile}
N.~{Chavan-Dafle} and A.~{Rodriguez}, ``Prehensile pushing: In-hand
  manipulation with push-primitives,'' in \emph{IEEE/RSJ International
  Conference on Intelligent Robots and Systems (IROS)}, 2015, pp. 6215--6222.

\bibitem{ma_toward_2017}
R.~R. {Ma}, W.~G. {Bircher}, and A.~M. {Dollar}, ``Toward robust, whole-hand
  caging manipulation with underactuated hands,'' in \emph{IEEE International
  Conference on Robotics and Automation (ICRA)}, 2017, pp. 1336--1342.

\bibitem{mason_push}
M.~T. {Mason}, ``Mechanics and planning of manipulator pushing operations,''
  \emph{The International Journal of Robotics Research}, vol.~5, no.~3, pp.
  53--71, 1986.

\bibitem{controller_refinement}
R.~{Platt}, R.~{Burridge}, M.~{Diftler}, J.~{Graf}, M.~{Goza}, E.~{Huber}, and
  O.~{Brock}, ``Humanoid mobile manipulation using controller refinement,'' in
  \emph{IEEE-RAS International Conference on Humanoid Robots}, 2006, pp.
  94--101.

\bibitem{funnel_tedrake}
A.~{Majumdar} and R.~{Tedrake}, ``Funnel libraries for real-time robust
  feedback motion planning,'' \emph{The International Journal of Robotics
  Research}, vol.~36, no.~8, pp. 947--982, 2017.

\bibitem{eppner_planning}
C.~{Eppner} and O.~{Brock}, ``Planning grasp strategies that exploit
  environmental constraints,'' \emph{IEEE International Conference on Robotics
  and Automation (ICRA)}, vol. 2015, pp. 4947--4952, 2015.

\bibitem{grebenstein_hand_2012}
M.~{Grebenstein}, M.~{Chalon}, W.~{Friedl}, S.~{Haddadin}, T.~{Wimböck},
  G.~{Hirzinger}, and R.~{Siegwart}, ``The hand of the {DLR} hand arm system:
  Designed for interaction,'' \emph{The International Journal of Robotics
  Research}, vol.~31, no.~13, pp. 1531--1555, 2012.

\bibitem{butterfass_dlr-hand_2001}
J.~{Butterfass}, M.~{Grebenstein}, H.~{Liu}, and G.~{Hirzinger}, ``{DLR}-hand
  {II}: next generation of a dextrous robot hand,'' in \emph{{IEEE}
  International Conference on Robotics and Automation (ICRA)}, vol.~1, 2001,
  pp. 109--114.

\bibitem{pfanne_object-level_2020}
M.~{Pfanne}, M.~{Chalon}, F.~{Stulp}, H.~{Ritter}, and A.~{Albu-Schäffer},
  ``Object-level impedance control for dexterous in-hand manipulation,''
  \emph{IEEE Robotics and Automation Letters}, vol.~5, no.~2, pp. 2987--2994,
  2020.

\bibitem{sundaralingam_relaxed-rigidity_2019}
B.~{Sundaralingam} and T.~{Hermans}, ``Relaxed-rigidity constraints: kinematic
  trajectory optimization and collision avoidance for in-grasp manipulation,''
  \emph{Autonomous Robots}, vol.~43, no.~2, pp. 469--483, 2019.

\bibitem{michelman_forming_1994}
P.~{Michelman} and P.~{Allen}, ``Forming complex dextrous manipulations from
  task primitives,'' in \emph{IEEE International Conference on Robotics and
  Automation (ICRA)}, vol.~4, 1994, pp. 3383--3388.

\bibitem{or_position-force_2016}
K.~{Or}, M.~{Tomura}, A.~{Schmitz}, S.~{Funabashi}, and S.~{Sugano},
  ``Position-force combination control with passive flexibility for versatile
  in-hand manipulation based on posture interpolation,'' in \emph{IEEE/RSJ
  International Conference on Intelligent Robots and Systems (IROS)}, 2016, pp.
  2542--2547.

\bibitem{data_driven_grasp}
J.~{Bohg}, A.~{Morales}, T.~{Asfour}, and D.~{Kragic}, ``Data-driven grasp
  synthesis—a survey,'' \emph{IEEE Transactions on Robotics}, vol.~30, no.~2,
  pp. 289--309, 2014.

\bibitem{rajeswaran_learning_2018}
A.~{Rajeswaran}, V.~{Kumar}, A.~{Gupta}, G.~{Vezzani}, J.~{Schulman},
  E.~{Todorov}, and S.~{Levine}, ``Learning complex dexterous manipulation with
  deep reinforcement learning and demonstrations,'' in \emph{Proceedings of
  Robotics: Science and Systems}, 2018.

\bibitem{nagabandi_deep_2019}
A.~{Nagabandi}, K.~{Konolige}, S.~{Levine}, and V.~{Kumar}, ``Deep dynamics
  models for learning dexterous manipulation,'' in \emph{Conference on Robot
  Learning (CoRL)}, 2019.

\bibitem{li_learning_2014}
M.~Li, H.~Yin, K.~Tahara, and A.~Billard, ``Learning object-level impedance
  control for robust grasping and dexterous manipulation,'' in \emph{{IEEE}
  International Conference on Robotics and Automation (ICRA)}, 2014, pp.
  6784--6791.

\bibitem{veiga_hierarchical_2020}
F.~Veiga, R.~Akrour, and J.~Peters, ``Hierarchical tactile-based control
  decomposition of dexterous in-hand manipulation tasks,'' \emph{Frontiers in
  Robotics and AI}, vol.~7, no. article no. 521448, 2020.

\bibitem{solak_learning_2019}
G.~{Solak} and L.~{Jamone}, ``Learning by demonstration and robust control of
  dexterous in-hand robotic manipulation skills,'' in \emph{IEEE/RSJ
  International Conference on Intelligent Robots and Systems (IROS)}, 2019, pp.
  8246--8251.

\end{thebibliography}

\clearpage

\end{document}